\documentclass[10pt,twocolumn,letterpaper]{article}

\usepackage[pagenumbers]{iccv} 

\usepackage{times}
\usepackage{soul}
\usepackage{url}
\usepackage[utf8]{inputenc}
\usepackage{caption}
\usepackage{graphicx}
\usepackage{amsmath}
\usepackage{amsthm}
\usepackage{booktabs}
\usepackage{algorithm}
\usepackage{algorithmic}
\usepackage[switch]{lineno}

\usepackage{multirow}
\usepackage{adjustbox} 
\usepackage{subcaption} 
\usepackage{amssymb}

\graphicspath{{figures/}}

\usepackage{gensymb}
\usepackage{makecell}
\DeclareMathOperator{\lat}{lat}
\DeclareMathOperator{\vecop}{vec}
\DeclareMathOperator{\unvecop}{unvec}
\DeclareMathOperator{\emb}{emb}

\usepackage{xspace}
\makeatletter
\DeclareRobustCommand\onedot{\futurelet\@let@token\@onedot}
\def\@onedot{\ifx\@let@token.\else.\null\fi\xspace}
\def\eg{\emph{e.g}\onedot} 
\def\ie{\emph{i.e}\onedot}

\makeatother

\newcommand{\RA}{\textit{CAMSRA}}
\newcommand{\FC}{\textit{CAMS Analysis}}

\definecolor{iccvblue}{rgb}{0.21,0.49,0.74}
\usepackage[pagebackref,breaklinks,colorlinks,allcolors=iccvblue]{hyperref}

\def\paperID{9663} 
\def\confName{ICCV}
\def\confYear{2025}

\title{Offline Meteorology-Pollution Coupling Global Air Pollution Forecasting Model with Bilinear Pooling}

\author{
Xu Fan\textsuperscript{1}\thanks{Equal contribution} \quad
Yuetan Lin\textsuperscript{1}\footnotemark[1] \quad
Bing Gong\textsuperscript{2} \quad 
Hao Li\textsuperscript{3,1}\thanks{Corresponding author} \\ 
{
\textsuperscript{1}{Shanghai Academy of AI for Science} \quad 
\textsuperscript{2} {Shanghai Normal University} \quad
\textsuperscript{3} {Fudan University}
}  \\
{\tt\small 
\{fanxu,linyuetan\}@sais.com.cn \quad
gongbing1112@gmail.com \quad
lihao\_lh@fudan.edu.cn
}
}

\begin{document}
\maketitle
\begin{abstract}
Air pollution has become a major threat to human health, making accurate forecasting crucial for pollution control.
Traditional physics-based models forecast global air pollution by coupling meteorology and pollution processes, using either online or offline methods depending on whether fully integrated with meteorological models and run simultaneously. However, the high computational demands of both methods severely limit real-time prediction efficiency.
Existing deep learning (DL) solutions employ online coupling strategies for global air pollution forecasting, which finetune pollution forecasting based on pretrained atmospheric models, requiring substantial training resources.
This study pioneers a DL-based offline coupling framework that utilizes bilinear pooling to achieve offline coupling between meteorological fields and pollutants.
The proposed model requires only 13\% of the parameters of DL-based online coupling models while achieving competitive performance.
Compared with the state-of-the-art global air pollution forecasting model CAMS, our approach demonstrates superiority in 63\% variables across all forecast time steps and 85\% variables in predictions exceeding 48 hours.
This work pioneers experimental validation of the effectiveness of meteorological fields in DL-based global air pollution forecasting, demonstrating that offline coupling meteorological fields with pollutants can achieve a 15\% relative reduction in RMSE across all pollution variables.
The research establishes a new paradigm for real-time global air pollution warning systems and delivers critical technical support for developing more efficient and comprehensive AI-powered global atmospheric forecasting frameworks.
\end{abstract}

\section{Introduction}
The rapid expansion of human productivity, coupled with accelerated industrialization, urbanization, and the global dependence on fossil fuels, has led to a significant increase in pollutant emissions~\cite{zhan2023impacts}. These emissions undergo complex chemical reactions and atmospheric transport processes, leading to severe air pollution, which causes approximately 6.7 million premature deaths annually~\cite{who2024report}. According to the World Health Organization (WHO), major pollutants in the atmosphere include particulate matter (PM) with diameters less than 10 micrometers (PM\textsubscript{10}) and 2.5 micrometers (PM\textsubscript{2.5}), sulfur dioxide (SO\textsubscript{2}), carbon monoxide (CO), ozone (O\textsubscript{3}), and nitrogen dioxide (NO\textsubscript{2})~\cite{who2024report}. In response to this public health and environmental crisis, governments worldwide implement proactive measures through development and enforcement of stringent environmental regulations~\cite{liao2018public,blackman2018efficacy}. Advancements in global air pollution forecasting systems enhance the capacity to provide early warnings for pollution events, reduce exposure risks, and support regional pollution control, policy development, and climate change mitigation, contributing to sustainable development~\cite{subramaniam2022artificial}.

Air pollution forecasting is a critical part of the global atmospheric forecasting system, where meteorological fields play a crucial role in the formation, transport, and transformation of pollutants~\cite{liu2020exploring,benarie1988applications}. Traditional numerical forecasting methods couple meteorological forecasting with air pollutants to simulate pollutants transport dynamics, and chemical transformations on a global scale. Recent advances in DL-based weather forecasting domain have significantly improved performance of global meteorological forecasting. Models such as FourCastNet~\cite{pathak2022fourcastnet}, Pangu-Weather~\cite{bi2022pangu}, GraphCast~\cite{lam2022graphcast}, FuXi~\cite{chen2023fuxi,sun2024fuxi} and AIFS~\cite{lang2024aifs} achieve competitive forecasting precision rivaling that of the high-resolution forecasts (HRES) of the European Centre for Medium-Range Weather Forecasts (ECMWF), one of the world's leading weather prediction systems. 
These advancements provide a strong foundation for developing an AI-powered global atmospheric forecasting system. However, research on DL-based global air pollutant forecasting remains limited. Using DL-based methods to couple air pollution forecasting with meteorological fields can overcome the constraints of traditional physical parameterization approaches, enhancing the efficiency and accuracy of air pollution predictions~\cite{bodnar2024aurora}. 

The meteorology-pollution coupling of traditional air pollution forecasting models can be classified into online and offline methods~\cite{martin2022improved}. For online methods, the air pollution model is integrated into the meteorological model, solving the continuity equations alongside the atmospheric dynamical equations~\cite{khan2021development}. 
In contrast, the offline methods treat the pollution model as an independent module, using external meteorological data (\eg, reanalysis or numerical weather prediction data) as input to solve the continuity equations~\cite{chen2021persistence}. Compared to online methods, offline methods are more computationally efficient and easier to deploy. 
However, both methods require meteorological forecasts for coupling, further increasing the computational burden of already expensive numerical weather prediction models and necessitating substantial resources for both development and operation.
DL-based models improve both forecasting efficiency and accuracy. Aurora~\cite{bodnar2024aurora} fine-tunes a foundation model pre-trained on meteorological data using the Copernicus Atmosphere Monitoring Service (CAMS) dataset~\cite{eskes2024evaluation}, the longest publicly available global air pollution dataset, enabling deep learning-based online coupled forecasting. However, its accuracy depends on the pre-training process, and online coupled fine-tuning is computationally expensive.

The DL-based offline approach for global air pollution forecasting demonstrates enhanced flexibility and adaptability to diverse meteorological field inputs, while simultaneously imposing a high demand for the intergration of meteorological and pollution data.
Bilinear pooling enables effective multimodal feature integration through high-order interactions that approximates vector outer products~\cite{lin2015bilinear,gao2016compact}.
By developing a low-rank approximation scheme combined with expanded receptive fields, we achieve efficient fusion of meteorological field and pollutant features.
Furthermore, the bilinear fusion inherently enables low-cost adaptation to different meteorological input configurations.

In this study, we present a novel offline-coupling AI framework for global air pollution forecasting which integrates meteorological fields and pollutants. The framework employs separate processing of meteorological and pollutant data streams, and designs an offline coupling module based on bilinear pooling, which collaboratively fuses multi-modal data. This architecture enhances computational efficiency while improving adaptability to heterogeneous meteorological forecasting models and enabling flexible cross-platform deployment. Additionally, we demonstrate the role of meteorological fields in global air pollution forecasting. These advancements provide both methodological innovations and technical support for developing AI-powered atmospheric forecasting systems.

The contributions can be summarized as follows:
\begin{itemize}
\item We propose an DL-based offline meteorology-pollution coupling framework that achieves competitive results with only 13\% of model parameters compared with online methods. We quantitatively validate the contribution of meteorological fields to air pollution prediction.
\item For meteorological and pollutant dual-modal data, we introduce deep integration via bilinear pooling, innovatively establishing nonlinear correlations between meteorological elements and pollutant dispersion in feature space.
\end{itemize}

\section{Related Works}
\textbf{Traditional physics-based models. }
Traditional physics-based models conduct global forecasts by solving a set of 3D continuity equations that describe emissions, transport, chemical reactions, and deposition~\cite{brasseur2017modeling}. Only a few models are capable of global air pollution forecasting. According to the World Meteorological Organization (WMO) Global to Local Air Quality Forecast Inventory~\cite{who2024Inventory}, the primary models offering real-time global air pollution forecasts include: the National Aeronautics and Space Administration's (NASA) GEOS-CF system, the National Center for Atmospheric Research's (NCAR) Community Earth System Model Version 2 (CESM2), the Finnish Meteorological Institute's SILAM, and the CAMS system~\cite{eskes2024evaluation}. These models offer global air pollution forecasts with spatial resolutions ranging from 0.25° to 1° and forecast lead times of 4 to 10 days. In this study, CAMS's data are selected for training and forecast performance comparison. CAMS system is operated by ECMWF, extends the atmospheric composition modeling capabilities of the Integrated Forecasting System (IFS), one of the world’s leading operational medium-range weather forecasting systems. CAMS employs an online coupling approach, integrating chemical and meteorological processes to generate seamless global forecasts. It produces forecasts twice daily at 00:00 and 12:00 UTC, providing five-day predictions at a spatial resolution of approximately $0.4^\circ \times 0.4^\circ$. Additionally, it offers publicly accessible datasets, including reanalysis, analysis, and forecast products spanning extended periods. To improve forecast accuracy, CAMS continuously updates its data assimilation and modeling systems in response to advancements in atmospheric observations and numerical techniques~\cite{peuch2022copernicus}. Quarterly evaluation reports assess forecast performance, ensuring global prediction reliability and enhancing the detection of major pollution events.

\textbf{DL-based models. }
The Aurora model~\cite{bodnar2024aurora} currently represents the first end-to-end DL architecture for global air pollution forecasting, employing a two-phase hierarchical training strategy: multi-scale pretraining on meteorological data and domain-specific fine-tuning with CAMS pollutant datasets. Computational constraints necessitate a batch size of 1 during both training phases. The foundation model undergoes 150,000 optimization steps across 32 GPUs. Subsequent fine-tuning comprises two stages: initial fine-tuning with CAMS reanalysis (35k steps) followed by analysis data adaptation (15k steps), totaling 50,000 steps. This meteorology-pollution coupling forecasting system demonstrates three critical characteristics: (a) tight coupling between meteorological and air pollutants processes through shared latent representations; (b) strong dependence on foundation model's weights for air pollutants finetuning, and (c) substantial computational requirements. Moreover, to ensure the effectiveness of the online coupling forecasting model, meteorological data must be consistent for both training and inference stages, which reduces the flexibility of model deployment. Our proposed offline coupling framework decouples meteorological dependencies, significantly reducing data dependency and training costs while achieving competitive forecasting performance.

\section{Methods}\label{sec_method}
\subsection{Preliminaries} 
Meteorological fields are pivotal in the transport, dispersion, transformation, and removal of air pollutants~\cite{liu2020exploring}. Wind speed and direction are key determinants of pollutants' horizontal transport, while atmospheric stratification directly affects their vertical dispersion. Furthermore, these conditions substantially regulate the chemical transformation of pollutants. Solar radiation and ambient temperatures catalyze photochemical reactions, and humidity influences heterogeneous reactions on aerosol surfaces, thus promoting the formation of secondary particulates. Additionally, meteorological variables govern the efficacy of pollutants removal via dry and wet deposition processes.

\subsection{Overview}
Unlike online approaches~\cite{bodnar2024aurora}, which rely on relatively large meteorological forecasting models and utilize pollutant data to fine-tune and update the model parameters for air pollution forecasting, our offline model is significantly smaller in scale. We directly forecast the future pollutant concentrations taking meteorological and pollutant data as input. Specifically, we input the pollutant concentrations at the initial time ($t$) and the previous time step ($t-1$), as well as the meteorological data at the initial time ($t$) and the next time step ($t+1$). By learning their changes and interaction relationships, we output the pollutant concentrations at the next time step. Since we are using an offline approach, the freedom of input allows us to explore the guiding role of meteorological elements in pollution forecasting.

\begin{figure*}[!t]%
\centering
\includegraphics[scale=0.8]{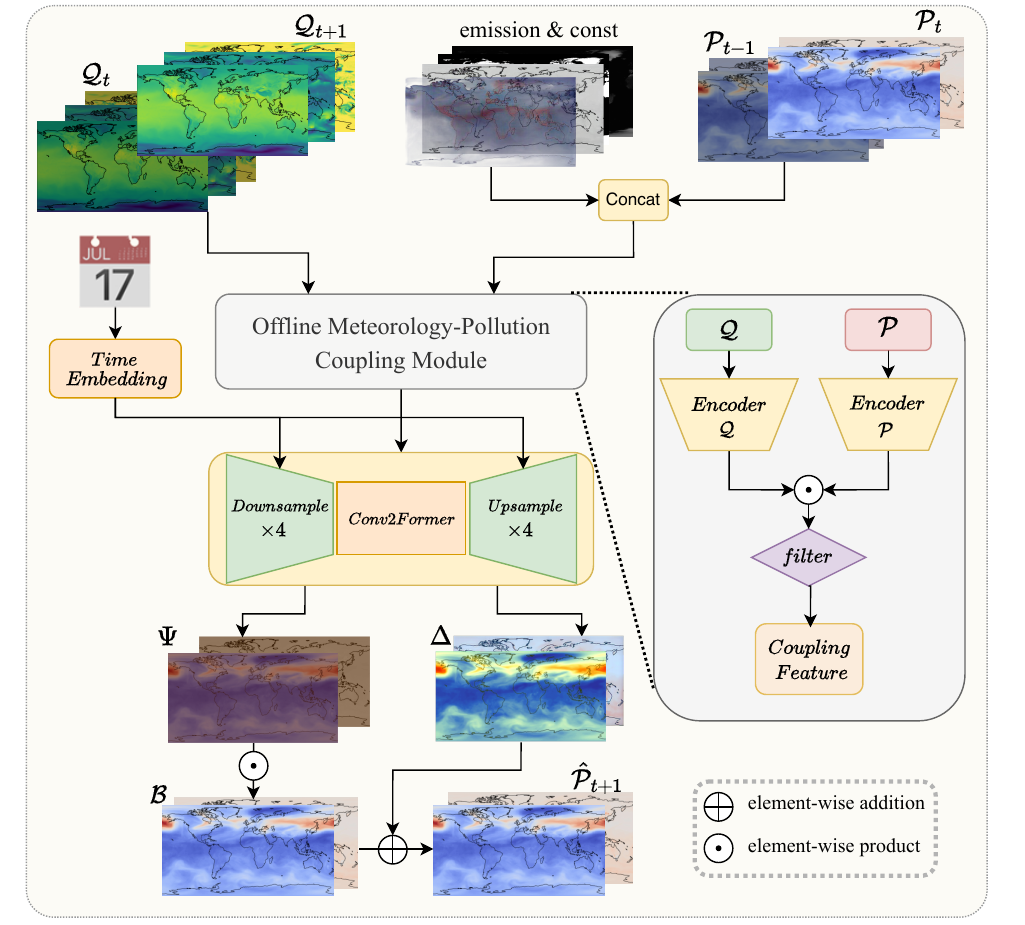}
\caption{Framework of our global air pollution forecasting.}\label{fig:framework}
\end{figure*}

As shown in Figure~\ref{fig:framework}, the pollutant concentrations ($\mathcal{P}$) and meteorological features ($\mathcal{Q}$) are represented in the form of tensors, the dimensions of which are $M\times N\times C$ and $M\times N\times D$, respectively, where $M$ and $N$ are the latitude and longitude sizes, $C$ and $D$ are the dimensions of the channels of pollutant and meteorological features, respectively.
The dimension $C$ of the pollutant features includes the concentrations of total column (TC) and 13 atmospheric levels of 5 chemical species (\ie, CO, NO, NO\textsubscript{2}, SO\textsubscript{2} and O\textsubscript{3}), as well as the concentrations of 3 particulate matter variables (\ie, PM\textsubscript{1}, PM\textsubscript{2.5} and PM\textsubscript{10}).
We concatenate the features at different time steps along the channel dimension.
Since factors such as emission data are involved in chemical reactions and factors like terrain affect the diffusion of pollutants, we take this kind of data together with the pollutant features as inputs.
The pollutant and meteorological features are fused and input into the model backbone. The time embedding, as a condition, also participates in the calculation of the backbone. The output results exert an influence on the pollutant base, and ultimately, the predicted pollutant concentration for the next time step is given.

\subsection{Changes Prediction}
To capture the changing trends of air pollutants, we take pollutant data at the initial time ($t$) and the previous time-step in history ($t-1$) as input and predict pollutant data at the next time-step ($t+1$).
Each time-step is 12 hours, which is sufficient for the vast majority of applications.
Specifically, we learn historical changes through pollutant data of $\mathcal{P}^{t}$ and $\mathcal{P}^{t-1}$, and meanwhile we use the forecasted meteorological data of $\mathcal{Q}^{t}$ and $\mathcal{Q}^{t+1}$ to predict future changes of pollutants. Instead of directly predicting the pollutant data at $(t+1)$, we predict the pollutant changes $\Psi$ and $\Delta$.
\begin{equation}
\Psi, \Delta=f_\theta(\mathcal{P}^{t-1}, \mathcal{P}^{t}, \mathcal{Q}^{t}, \mathcal{Q}^{t+1}), \label{eq1}
\end{equation}
where $\Psi$ is the scaling factor and $\Delta$ is the amount of change, $\theta$ is the network parameters.
The pollutant output ($\hat{\mathcal{P}}$) is obtained by adding changes to the pollutant base ($\mathcal{B}$), as demonstrated in Eq.~\ref{eq2}.
\begin{equation}
\hat{\mathcal{P}}=\Psi\times\mathcal{B}+\Delta.\label{eq2}
\end{equation}

\subsection{Pollutant Base Construction}
The principle for building the pollutant base is that the pollutant data should not be after the initial time and should be as close as possible to the data at the forecast time. Empirically, we have found that some pollutant variables that are prone to react under sunlight have the problem of the grey line\footnote{A shifting line around the Earth that separates daylight and darkness parts.} (also referred to as the terminator or the twilight zone).
That is, the global distribution of these pollutants changes along with the grey line.
Therefore, when constructing the pollutant base, for the variables with the grey line issue, we select the data 24 hours before the forecast time, and for other variables, we select the most recent data (the data at the initial time).
To predict more time steps, we use the already predicted data as input and perform cyclic prediction of the subsequent steps.

\subsection{Model Structure}
Benefiting from the symmetrical encoder-decoder model structure, U-Net~\cite{ronneberger2015u} can learn different semantic features at different levels, and thus we adopt U-Net as our backbone network.
The encoder module consists of 4 layers that down-sample the input features layer by layer and simultaneously increase the number of channels to enhance the semantic representation.
The encoder module achieves the compression of the input features and the intermediate output features are restored to the original feature size through the four layers of upsampling in the decoder module.
Meanwhile, through skip-connections, U-Net fuses the shadow detailed information such as texture and position from the encoder path, as well as the deep semantic information from the decoder path, and can better accelerate convergence.
For the hidden layers between the encoder and the decoder, we adopt the Conv2Former~\cite{hou2024conv2former} module.
We embed the time variable variable and add them at each layer of U-Net to provide additional information. 

\subsection{Efficient Feature Fusion}
U-Net takes only one tensor as input.
In order to fuse the pollutant features ($\mathcal{P}_{in}=[\mathcal{P}^{t-1}, \mathcal{P}^{t}]$) and meteorological features ($\mathcal{Q}_{in}=[\mathcal{Q}^{t}, \mathcal{Q}^{t+1}]$) as input, a straightforward solution is to concatenate them along the channel dimension ($\mathcal{X}=[\mathcal{P}_{in}, Q_{in}]$).
The dimensions of $\mathcal{P}_{in}$ and $\mathcal{Q}_{in}$ are $M\times N\times 2C$ and $M\times N\times 2D$, respectively.
In addition to simple concatenation of features, bilinear pooling provides a richer information interaction for the fusion of features of two modalities~\cite{lin2015bilinear}.
However, the dimension of the fused features is too high, which puts great pressure on the efficiency of model training, and meanwhile, is prone to lead to the overfitting problem.
Some works perform a low-rank approximation on the parameter matrix of the bilinear model after decomposition~\cite{gao2016compact}, achieving the goal of improving both efficiency and effectiveness.

We achieve low-rank bilinear pooling through an approximation scheme that performs the Hadamard product on the two modal features after mapping them to a common space~\cite{kim2022hadamard}.
Specifically, given the pollutant and meteorological features at location $(i, j)$, we denote the pollutant and meteorological vectors as $\boldsymbol{p}_{ij}\in\mathbb{R}^{2C}$ and $\boldsymbol{q}_{ij}\in\mathbb{R}^{2D}$.
The fused feature $\boldsymbol{x}_{ij}$ is shown below.
\begin{equation}
\boldsymbol{x}_{ij}=W_x^T(W_p^T\boldsymbol{p}_{ij}\circ W_q^T\boldsymbol{q}_{ij}), \label{eq3}
\end{equation}
where $W_p\in\mathbb{R}^{2C\times H}$, $W_q\in\mathbb{R}^{2D\times H}$ and $W_x\in\mathbb{R}^{H\times G}$ are parameters of the mapping layer.
The hyperparameter $H$ is the number of channels in the hidden layer and $G$ is the size of the input feature of U-Net.
Meteorological factors and pollutants are greatly influenced by adjacent areas.
Therefore, we increased the receptive field and the experimental results have shown that the predictive ability of the model is significantly improved after increasing the receptive field.
Eq.~\ref{eq3} can be expressed as the following formula.
\begin{equation}
\boldsymbol{x}_{mn}=\unvecop({W'}_x^T({W'}_p^T\vecop(\boldsymbol{p}_{mn})\circ {W'}_q^T\vecop(\boldsymbol{q}_{mn}))), \label{eq4}
\end{equation}
where $m$ and $n$ denote location range $[i-1:i+1]$ and $[j-1:j+1]$, $\vecop$ and $\unvecop$ are the process of flattening the tensor into a vector and the inverse process, respectively.
Correspondingly, the size of the parameters of the mapping layer ${W'}_p$, ${W'}_q$ and ${W'}_x$ becomes $(2C*9)\times H$, $(2D*9)\times H$ and $H\times 9G$.
The fused features are input to the U-Net model, and after being processed by the encoder-decoder, the pollutant changes are given.

\section{Experiments}\label{sec_exp}

\subsection{Datasets}
We use meteorological and pollutant data in this study.
The meteorological data includes ERA5 reanalysis data~\cite{hersbach2020era5} used in the training phase and FuXi~\cite{chen2023fuxi,sun2024fuxi} forecasting data used in the inference phase.
The meteorological data is linearly interpolated to a resolution of 0.4\degree~to match the spatial resolution of CAMS analysis and forecast fields, thus giving $M=451$ and $N=900$.

We normalize meteorological input and output data according to FuXi model's standardization scheme.
We also devise a tailored normalization approach for pollutant data to account for their skewed distributions.
We design an extra transformation function to mitigate the impact of specific pollutant variables (\eg, NO\textsubscript{2} and SO\textsubscript{2}) due to sharp spike distribution in regions with dense human activities.
This gives a rapid boost to smaller values, enhancing the sensitivity to low-magnitude values.
For more details of the dataset, please refer to the supplementary materials.

\subsection{Setup}
\textbf{Implementation Details. } 
We adopt a two-stage training scheme.
We pretrain our base model using long-range {\RA} and fine-tune the model on relatively more accurate {\FC}.
We implement our models in PyTorch and train them using one NVIDIA A100 GPU card.
During the model pretraining stage, we use 9 years of {\RA} data and train for 40 epochs.
For the model fine-tuning stage, 1.5 years of {\FC} data is used.
We adopt a relatively small learning rate and train 100 epochs for single-step mode and 20 epochs for multi-step expansion.

We schedule the learning rate (lr) as follows.
First, we use $1/3$ of the first epoch for warm-up, during which lr increases linearly from 1e-8 to the maximum lr.
Then, we use the cosine annealing method to gradually decrease lr to 1e-9 for the remaining training iterations.
The maximum lr is set to 2.5e-4 for pretrain stage and 1e-5 for fine-tuning stage.

\textbf{Time embedding. }
Since pollutant concentrations exhibit a distinct diurnal cycle, we encode time using a Fourier encoding scheme. Specifically, the initial time is converted into day of the year and hour of the day, with corresponding wavelengths ($\lambda$) set to 366 and 24, respectively.
\begin{align}
\emb(t)=[\cos(\frac{t}{\lambda}), \sin(\frac{t}{\lambda})]
\end{align}

\textbf{Loss Setting. }
Due to the spherical shape of the Earth, we assign weights that decrease as latitude increases.
The weights $\boldsymbol{w}\in\mathbb{R}^M$ is computed as below, 
\begin{equation}
\boldsymbol{w}_i=\cos(\lat(i)), \label{eq_wgt}
\end{equation}
where $\lat(\cdot)$ is the conversion function from latitude index $i\in[0,450]$ to radian, and $\boldsymbol{w}_i$ progressively increases from near-zero value to approach one before diminishing back to near-zero value.
Based on the pollutant base and pollutant forecast groundtruth, we calculate the weighted average error $\boldsymbol{e}\in\mathbb{R}^C$, which shows the range of pollutant variation from baseline to ground-truth values, indicating how easily they can change.
\begin{align}
\boldsymbol{e}_k=&\sum_{i, j}\left|\boldsymbol{b}_{ijk}-\boldsymbol{p}_{ijk}\right|\boldsymbol{w}_i,
\label{eq:err}
\end{align}
where $k$ is the index of pollutant variable.
Similarly, when calculating the loss, we compute the error between the pollutant forecast output and the ground-truth, and regulate the differences of different variables via the weighted error in Eq.~\ref{eq:err}.
\begin{align}
\hat{\boldsymbol{e}}_k=&\sum_{i, j}\left|\hat{\boldsymbol{p}}_{ijk}-\boldsymbol{p}_{ijk}\right|\boldsymbol{w}_i, \\
loss=&\frac{1}{C}\sum_k\frac{\hat{\boldsymbol{e}}_k}{\boldsymbol{e}_k}=\frac{1}{C}\sum_k\frac{\sum_{i, j}\left|\hat{\boldsymbol{p}}_{ijk}-\boldsymbol{p}_{ijk}\right|\boldsymbol{w}_i}{\sum_{i, j}\left|\boldsymbol{b}_{ijk}-\boldsymbol{p}_{ijk}\right|\boldsymbol{w}_i}.
\end{align}

\begin{figure*}[!t]%
\centering
\includegraphics[scale=0.55]{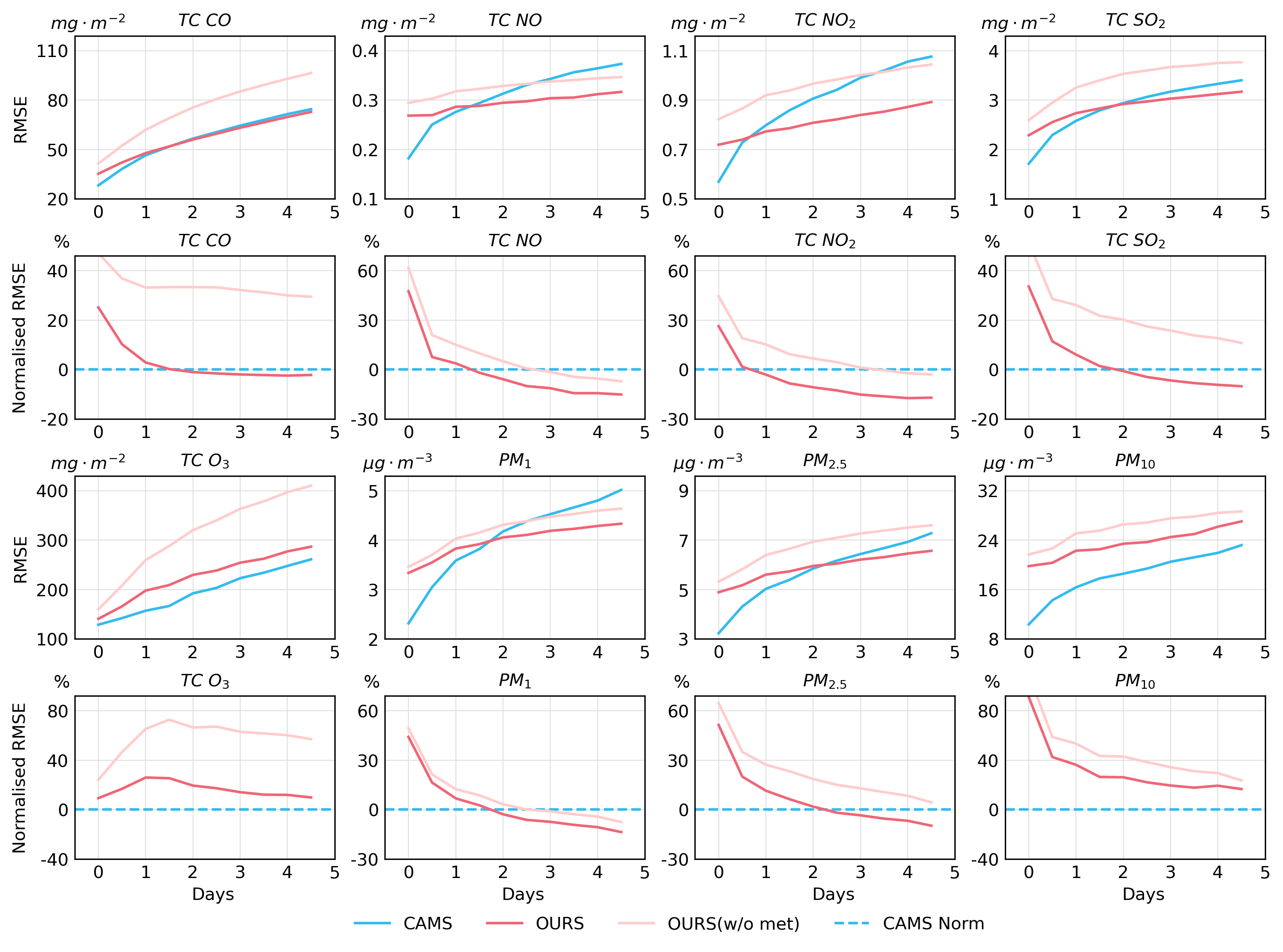}
\caption{Weighted root mean square error (RMSE) comparison of CAMS model v.s. ours on pollutant variables of total column (TC) and particulate matter (PM). Normalized RMSE values based on CAMS results are given below each RMSE curves. (Best viewed in color.)}\label{fig_rmse}
\end{figure*}

\subsection{Experimental Results}
Given the lack of open-source air pollution forecasting models, we conduct comparative evaluations of our framework against alternative configurations and the physics-based CAMS model.
The evaluations utilize {\FC} data from June to November 2022 (inclusive), a period characterized by relatively stable conditions including minimal emission inventory changes, fixed pollutant types, and consistent chemical mechanisms, which are critical prerequisites to ensure evaluation validity.
We demonstrate the results as follows.

\textbf{Surface-level Variables. } Fig.~\ref{fig_rmse} compares the weighted root mean square error (RMSE) and normalized RMSE value curves of the CAMS model and our model (OURS) on the pollutant variables of total column (TC) (\ie, CO, NO, NO\textsubscript{2}, SO\textsubscript{2} and O\textsubscript{3}), and variables of particulate matter (PM) (\ie, PM\textsubscript{1}, PM\textsubscript{2.5} and PM\textsubscript{10}). The RMSE of all three models gradually increased over time, with the variations of our two models stabilizing after 48 hours. However, CAMS exhibited a more pronounced RMSE increase after 48 hours, whereas OURS demonstrated a monotonic decrease in Normalized RMSE, particularly during the first 48 hours. A comparative analysis indicates that OURS outperforms CAMS in the later stages of forecasting, particularly for TC CO, TC NO\textsubscript{2}, TC NO, TC SO\textsubscript{2}, PM\textsubscript{1}, and PM\textsubscript{2.5}, highlighting its advantage in long-term predictions. 

The observed situation is attributed to the online and offline coupling methods of the models, and the accuracy of the meteorological forecast fields. A notable bidirectional feedback exists between air pollutants and meteorological conditions. For instance, PM\textsubscript{2.5} interacts with radiation, where local increases in PM\textsubscript{2.5} concentration absorb and scatter solar radiation, causing surface cooling that suppresses boundary layer development, thus exacerbating pollutant accumulation. While the CAMS system employs online coupling that theoretically captures this feedback mechanism through IFS-derived initial meteorological fields during the early forecast period. However, since CAMS also uses an autoregressive forecasting method, the errors of both meteorological and pollutant forecast fields gradually accumulate over time. In contrast, our model demonstrates superior meteorological forecasting performance compared to IFS, coupled with an offline coupling architecture which prevents the accumulation of errors from pollutant forecasts affecting meteorological fields. This configuration enables sustained prediction accuracy throughout extended forecast periods.

The absence of meteorological coupling in AI forecasting models leads to progressive performance degradation. Although initial forecasts show comparable accuracy to offline coupling systems, the divergence amplifies with extended forecast horizons. This indicates that the offline coupled meteorological fields help mitigate the cumulative errors in autoregressive forecasts. Notably, the optimization effect for O\textsubscript{3} is particularly remarkable, with a 32\% relative reduction in RMSE in all forecast time steps. This improvement is attributed to meteorological modulation of photochemical production, vertical mixing efficiency, and wet deposition processes, all of which are crucial for both short-term and long-term variations in O\textsubscript{3} concentrations. Similarly, TC SO\textsubscript{2} errors decrease by 18\% through improved characterization of dry deposition velocities and scavenging coefficients. TC CO exhibits relative chemical stability, and 26\% error reduction reflects enhanced transport process representation, particularly in boundary layer venting mechanisms. PM\textsubscript{2.5} forecasts show 14\% improvement as meteorological drivers govern both long-range transport and secondary aerosol formation. 

\begin{figure}[!t]%
\centering
\includegraphics[scale=0.40]{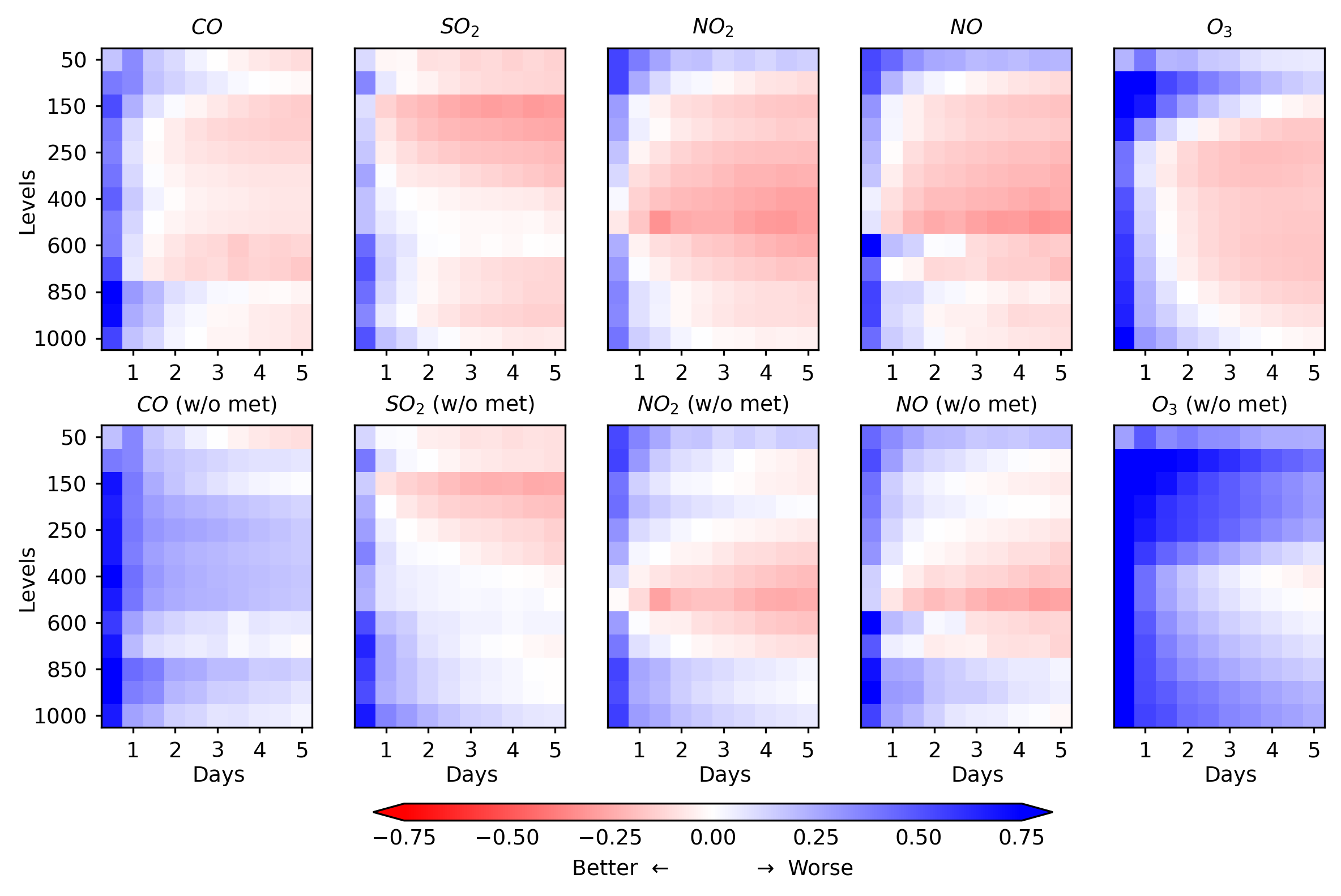}
\caption{The normalized RMSE score matrix relative to the CAMS model for the pollutant variables at all pressure levels and all lead days. Red represents good results. (Best viewed in color.)}\label{fig_scorecard}
\end{figure}

\begin{figure*}[!t]%
\centering
\includegraphics[scale=0.5]{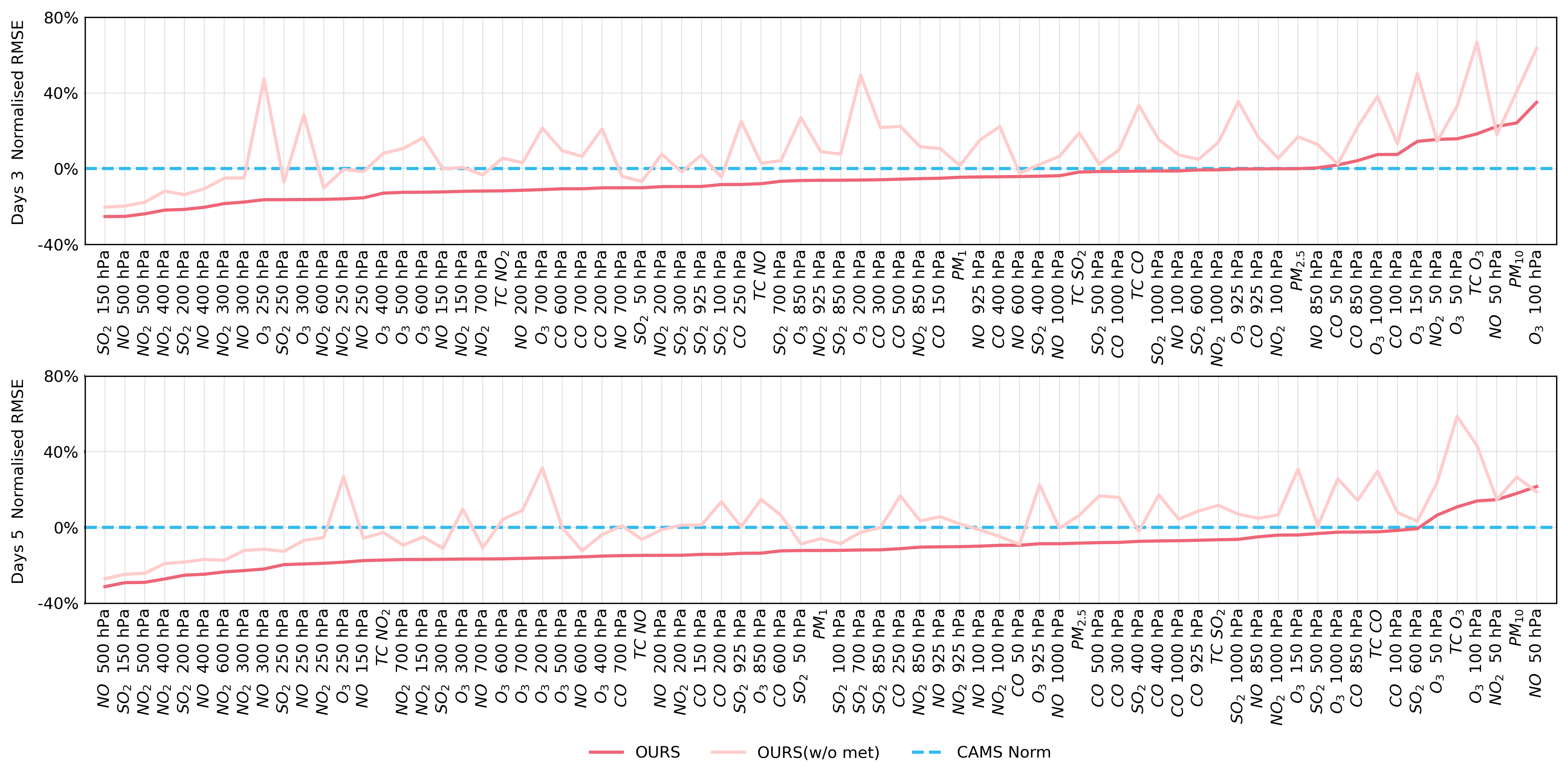}
\caption{The normalized RMSE values relative to the CAMS model at 3/5 days of lead time.
A lower normalized value indicated better performance.}\label{fig_3dleadtime}
\end{figure*}

\textbf{Pressure-level Variables. } The forecasting performance of OURS shows a consistent trend over time across different pressure levels, as in Fig.~\ref{fig_scorecard}. The vertical distribution shows that OURS performs significantly better above 850 hPa. The boundary layer is characterized by complex turbulent, thermodynamic, and material exchange processes, which increase forecast uncertainty. Traditional physical-based models incorporate extensive physical constraints in this region to improve forecast accuracy. In contrast, DL-based models implicitly learn from data about the various influencing factors and physical constraints within the boundary layer, resulting in limited advantages. However, in the middle and upper atmosphere, where pollutant concentrations are mainly governed by dynamical processes, DL-based models tend to outperform physical models. Without meteorological inputs, the forecast accuracy for CO and O\textsubscript{3}, which are highly sensitive to transport processes, deteriorates in the middle and upper atmosphere.

Based on the lifecycle and diffusion rates of pollutants, the key forecasting windows for pollution alerting and control are 12–72 hours (short-term) and 3–7 days (medium-term). We compute the average normalised RMSE, focusing on the second and fifth days within these critical time windows in Fig.~\ref{fig_3dleadtime}. Our model shows an advantage in only 6.8\% of variables in the 1-day forecast. However, this advantage increases to 49\% of variables in the 2-day forecast, and 91\% of variables outperform CAMS in the 5-day forecast. As the forecast duration increases, OURS's advantages become more pronounced, especially for variables in the middle and upper atmospheric pressure layers. However, forecast results for high-altitude layers (above 100 hPa) and PM\textsubscript{10} remain suboptimal. Notably, our model outperforms CAMS in 84\% of variables in the 3-day forecast, whereas the Aurora global air pollution model shows an advantage in 86\% of variables. However, Aurora's model contains 1.3B parameters, 7.3 times more than our model.

\begin{figure}[!t]%
\centering
\includegraphics[scale=0.46]{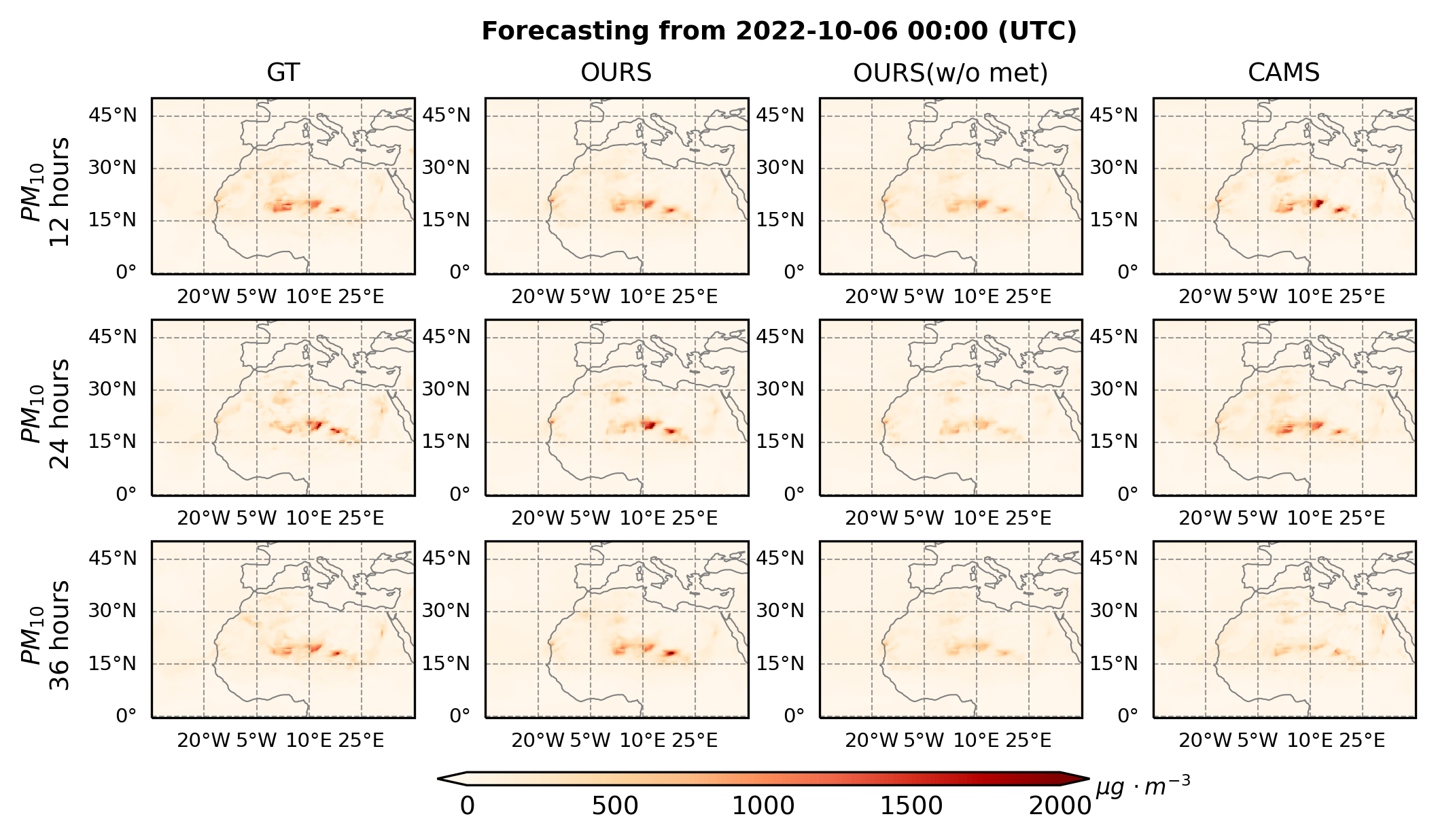}
\caption{Case study of PM\textsubscript{10} in North Africa, we list the forecast results for the next 12, 24 and 36 hours. The columns are respectively the groundtruths, predictions of our model and forecast results of CAMS. (Best viewed in color.)}\label{fig_case_pm10}
\end{figure}

\textbf{Cases Study. } In early October 2022, a significant dust event in the Sahara Desert sent plumes across the Atlantic, affecting the Caribbean Sea and southern Europe. Fig.~\ref{fig_case_pm10} displays various model forecasts. Initially, all models pinpoint high-value concentration centers. However, as forecasts extended, OURS (w/o met) and CAMS models show notable concentration decreases and shifts, indicating PM\textsubscript{10} dispersion and dilution. OURS maintains better consistency with the target, preserving the high-value center location. Both OURS and CAMS predict dust plume dispersion toward the Atlantic, though OURS (w/o met) has less pronounced dispersion. This underscores the importance of meteorological factors in pollutants transport forecasts. The proposed offline coupling mechanism effectively captures these drivers, surpassing traditional models in predicting PM\textsubscript{10} high-value centers during dust events.

\begin{figure}[!t]%
\centering
\includegraphics[scale=0.46]{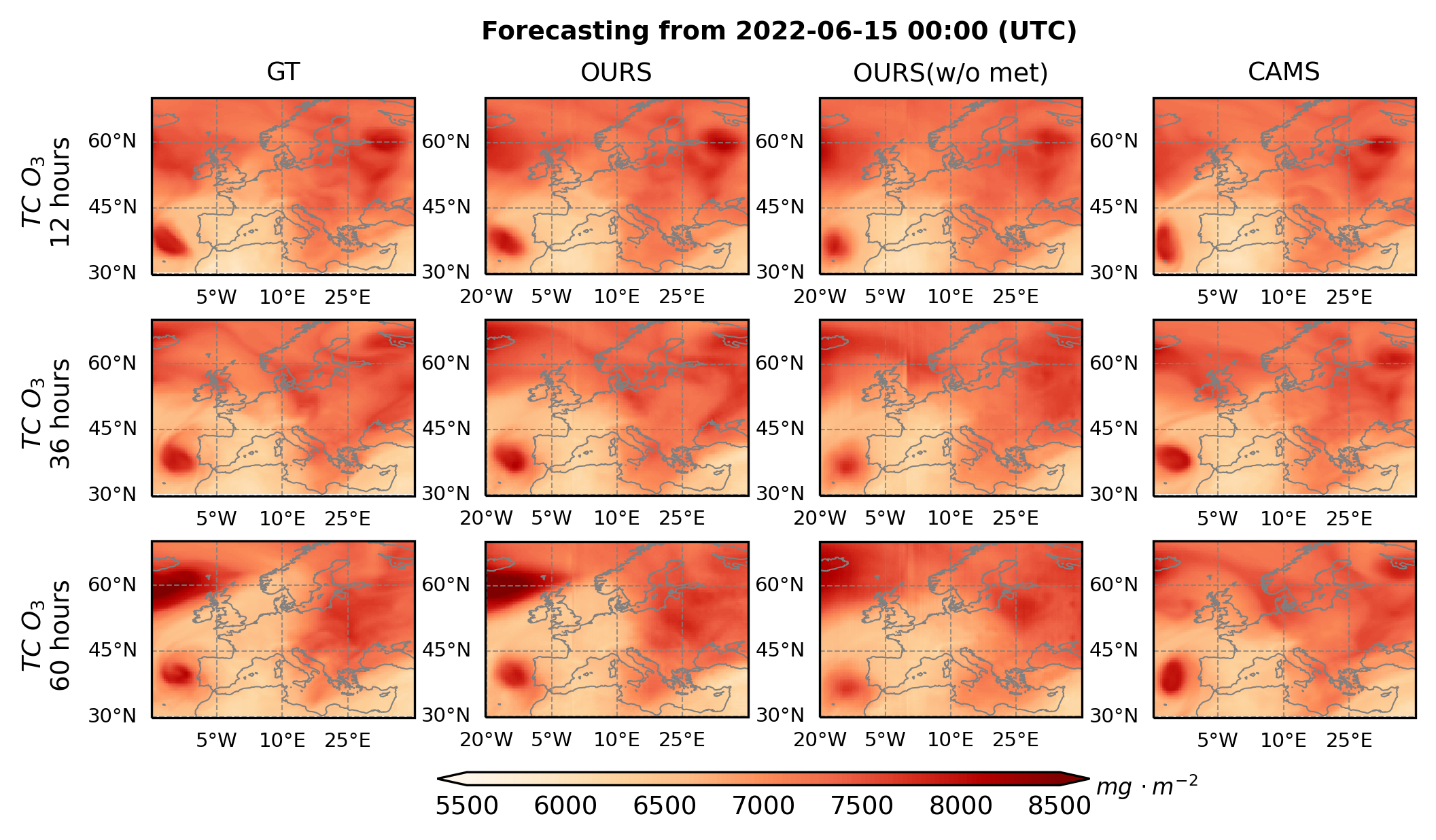}
\caption{Case study of TC O\textsubscript{3} in Europe, we list the forecast results for the next 12, 36 and 60 hours. The columns are respectively the groundtruths, predictions of our model and forecast results of CAMS. (Best viewed in color.)}\label{fig_case_gcto3}
\end{figure}

From June to September 2022, Europe experienced an extreme heatwave, with June temperatures $2.33^{\circ}C$ above the norm~\cite{ballester2023heat}. This intensified atmospheric photochemical reactions, leading to a rise in ozone concentrations, which typically increase after sunrise and peak in the afternoon. Considering the model's forecast resolution, Fig.~\ref{fig_case_gcto3} shows the TC O\textsubscript{3} distribution across Europe from June 15 at 00:00 UTC to June 17 at 12:00 UTC. Initially, OURS, OURS (w/o met), and CAMS models align closely in the 12-hour forecasts. However, as forecasts extended, OURS (w/o met) shows more uniformly distributed and lower concentration centers, significantly diverging. CAMS also differs markedly from target in northeastern Europe, notably failing to show dual TC O\textsubscript{3} concentration centers in the 60-hour forecast. Meanwhile, OURS consistently matches targets in multi-step forecasts, indicating that proposed offline coupling mechanism captures the spatiotemporal changes in TC O\textsubscript{3} concentrations under high temperatures more effectively than traditional models.

\section{Conclusion}
This study addresses the dual bottlenecks of high computational redundancy in traditional physical models and the large parameter sizes of existing deep learning online coupling frameworks. We innovatively propose a bilinear pooling-based meteorology-pollution offline coupling framework for global air pollution forecasting.
By establishing bilinear pooling based interaction mechanisms between meteorological fields and pollutants, our approach compresses model parameters to 13\% of online coupling solutions.
Experimental results validate the optimization potential of deep learning implicit coupling over traditional physics-based explicit modeling. This breakthrough not only lays an algorithmic foundation for building next-generation intelligent atmospheric monitoring systems but also holds significant practical implications for advancing digital response systems in global environmental governance.
Future research could achieve enhanced predictive capabilities through seamless integration with higher-precision pollution monitoring station data.

{
\small
\bibliographystyle{ieeenat_fullname}
\bibliography{main}
}

\appendix
\section{Datasets Introduction}\label{sec_dataset}

\textbf{Datasets. } This study utilized meteorological and atmospheric composition data, with the variables used listed in Table 1. The meteorological data include ERA5 reanalysis data and FuXi model forecast outputs, while the atmospheric composition data are sourced from CAMS. To align with FuXi's forecast results, 13 pressure levels consistent with FuXi were selected, and all data were linearly interpolated to a resolution of 0.4\degree~to match the spatial resolution of CAMS analysis and forecast fields. ERA5 data are used for coupling meteorological inputs during the training of the air pollutants forecasting model, while FuXi model forecast outputs replace ERA5 data during model testing.

\textbf{CAMS. } CAMS (Copernicus Atmosphere Monitoring Service) is an atmospheric composition analysis and forecasting system based on the ECMWF IFS model. It integrates modules for aerosols, reactive gases, and greenhouse gases, using a {4D}-Var data assimilation system to combine diverse observations (\eg, O\textsubscript{3}, CO, NO\textsubscript{2}, AOD with anthropogenic (MACCity), biogenic (MEGAN), and fire emissions (GFAS). CAMS provides global analysis, forecast, and reanalysis products.

CAMS Analysis and forecast Data: CAMS analysis data ({\FC}) are generated using ECMWF's 4D-Var assimilation method, integrating global meteorological, satellite, and ground-based observations to provide accurate analyses of aerosols, reactive gases, and greenhouse gases, which serve as initial conditions for global pollutant forecasting. CAMS forecast data are coupled with ECMWF’s IFS prediction system, combining meteorological forecasts with atmospheric composition simulations. Meteorological data from IFS provide boundary conditions for CAMS modules on aerosols and reactive gases, which interact with atmospheric chemical reaction modules to form feedback mechanisms for spatial-temporal predictions of atmospheric composition. CAMS forecasts cover a temporal range of one to five days with a typical spatial resolution of 0.4\degree, offer high-precision global atmospheric composition predictions. Given the frequent updates to CAMS forecasting models driven by advancements in pollutant forecasting and the incorporation of additional observational data, this study utilized CAMS forecast data from October 2020 to December 2023, a period of relative stability. 

CAMS Reanalysis data: CAMS reanalysis data ({\RA}) have a spatial resolution of 0.75\degree~with 60 vertical levels, covering 2003-2021 and is based on an older version of IFS (CY42R1). CAMSRA data exhibit good stability and superior precision in the analysis of ozone, CO, NO\textsubscript{2} and AOD compared to MACCRA and CIRA. However, due to its coarser resolution and older model cycle, CAMSRA is less precise than CAMS analysis and forecast data. In this study, CAMSRA data from 2010 to 2018 were used for pre-training.

\textbf{ERA5. } ERA5 is the fifth generation of ECMWF reanalysis datasets, providing a wealth of surface and upper-air variables. The dataset is generated by assimilating high-quality and abundant global observations using ECMWF's IFS model. It features a temporal resolution of 1 hour and a spatial resolution of 0.25\degree, covering data from January 1950 to the present. With its extensive temporal and spatial coverage and exceptional accuracy, ERA5 is widely recognized as one of the most comprehensive and precise reanalysis datasets globally.

\textbf{FuXi. } The FuXi weather model, developed based on the ECMWF ERA5 reanalysis dataset, employs a cascaded approach to provide 15-day global forecasts with a temporal resolution of 6 hours and a spatial resolution of 0.25\degree. FuXi achieves performance comparable to ECMWF's traditional numerical weather prediction model (EM) over 15-day forecasts while requiring significantly fewer computational resources. In this study, FuXi forecast data initialized at 00 and 12 UTC were used as meteorological inputs during the model testing phase, with a temporal resolution of 12 hours and a forecast range of up to 5 days.

\begin{table*}
\centering
\begin{tabular*}{\textwidth}{@{\extracolsep\fill}ccccc@{}}
\toprule
\textbf{Name} & \textbf{Resolution} & \textbf{Timeframe} & \textbf{Surface Variables} & \textbf{Atmospheric Variables} \\
\midrule
CMAS   & $0.4^\circ \times 0.4^\circ$ & 2010-2018   & \makecell[c]{TC CO, TC NO, TC NO\textsubscript{2}, TC SO\textsubscript{2}, \\ TC O\textsubscript{3}, PM\textsubscript{1}, PM\textsubscript{2.5}, PM\textsubscript{10}} & CO, NO, NO\textsubscript{2}, SO\textsubscript{2}, O\textsubscript{3} \\
CAMSRA & $0.75^\circ \times 0.75^\circ$ & Oct 2020-2023 & \makecell[c]{TC CO, TC NO, TC NO\textsubscript{2}, TC SO\textsubscript{2}, \\ TC O\textsubscript{3}, PM\textsubscript{1}, PM\textsubscript{2.5}, PM\textsubscript{10}} & CO, NO, NO\textsubscript{2}, SO\textsubscript{2}, O\textsubscript{3} \\
ERA5   & $0.25^\circ \times 0.25^\circ$ & 2010-2023   & T2M, U10M, V10M, MSL, TP   & U, V, T, RH, Z \\
FuXi   & $0.25^\circ \times 0.25^\circ$ & 2022-2023   & T2M, U10M, V10M, MSL, TP   & U, V, T, RH, Z \\
Geographical & $0.25^\circ \times 0.25^\circ$  & --          & orography, land-sea mask, latitude, longitude & -- \\
Temporal & --                       & --          & hour of day, day of year, step & -- \\
\bottomrule
\end{tabular*}
\caption{Summary of the datasets used to train and evaluate in this work.}
\label{tab_dwsc}
\end{table*}

\section{Datasets Normalization}\label{sec_datanorm}

Both inputs and outputs of the model are normalized, with meteorological variables normalized according to the FuXi model's standardization scheme. For pollutant data, a tailored normalization approach is devised to account for their skewed distributions.

For all air pollutant variables and emission inventory variables, we adopt the normalization method inspired by Aurora. Specifically, we estimate $scale_{v}$ as half the spatial maximum averaged over time. By construction of this normalization, the normalized air pollutant variables will typically be in the range [0, 2]. Normalization statistics are computed separately for {\RA} and {\FC} data.
\begin{align}
X_{v, i, j, norm}^{t}=\frac{X_{v, i, j}^{t}}{scale_v},
\end{align}
\begin{align}
scale_v=\frac{1}{2} \cdot\ \frac{1}{T} \sum_{t=1}^{T}max(x_{v, i, j}^t).
\end{align}

Concentration values of atmospheric pollutants, such as NO\textsubscript{2} and SO\textsubscript{2}, are closely related to human activities. These pollutants are primarily generated by human activities and then dispersed through atmospheric motion. As a result, their spatial distribution exhibits sharp spikes in areas with dense human activity, leading to a skewed overall distribution. Training directly on such data would make the network overly sensitive to high-magnitude values while being less responsive to low-magnitude values. To address this issue, we applied a transformation to normalized data. This transformation increases the variation in low-magnitude values while ensuring the transformed data remains monotonically increasing with respect to the original values. After extensive experimentation, we designed the following transformation function:
\begin{align}
x_{trans}=x_{norm}+\frac{log_{10}(x_{norm} \cdot 2.5\cdot10^4)}{log_{10}(25)}.\label{eq:sup3}
\end{align}

Eq.~\ref{eq:sup3} represents a monotonically increasing function. After normalization, air pollution variables will typically be in the range [0, 2]. When $x\leqslant 1\cdot e^{-2}$, $x_{trans}$  primarily depends on the second term, which enhances sensitivity to low-magnitude values. When  $x> 1\cdot e^{-2}$, the second term gradually becomes smoother, and the transformation is mainly influenced by the first term. This approach highlights the spatial distribution features of low-magnitude values, making the data more distinguishable.

\end{document}